\documentclass{article}

 \usepackage[preprint]{neurips_2025}

\usepackage[utf8]{inputenc} 
\usepackage[T1]{fontenc}    
\usepackage{hyperref}       
\usepackage{url}            
\usepackage{booktabs}       
\usepackage{amsfonts}       
\usepackage{nicefrac}       
\usepackage{microtype}      
\usepackage{xcolor}         
\usepackage{subcaption}
\usepackage{multirow}
\usepackage{bbm}
\usepackage{verbatim}
\usepackage[pdftex]{graphicx}

\title{A mini-batch training strategy for deep subspace clustering networks}

\author{%
  Yuxuan Jiang $^1$\thanks{E-mail: jiangyuxuan372@gmail.com} \quad Chenwei Yu $^1$ \quad Zhi Lin $^1$ \quad Xiaolan Liu $^2$\\
  $^1$The Hong Kong University of Science and Technology \\
  $^2$South China University of Technology \\
}

\begin{document}

\maketitle

\begin{abstract}

Mini-batch training is a cornerstone of modern deep learning, offering computational efficiency and scalability for training complex architectures. However, existing deep subspace clustering (DSC) methods, which typically combine an autoencoder with a self-expressive layer, rely on full-batch processing. The bottleneck arises from the self-expressive module, which requires representations of the entire dataset to construct a self-representation coefficient matrix. In this work, we introduce a mini-batch training strategy for DSC by integrating a memory bank that preserves global feature representations. Our approach enables scalable training of deep architectures for subspace clustering with high-resolution images, overcoming previous limitations. Additionally, to efficiently fine-tune large-scale pre-trained encoders for subspace clustering, we propose a decoder-free framework that leverages contrastive learning instead of autoencoding for representation learning. This design not only eliminates the computational overhead of decoder training but also provides competitive performance. Extensive experiments demonstrate that our approach not only achieves performance comparable to full-batch methods, but outperforms other state-of-the-art subspace clustering methods on the COIL100 and ORL datasets by fine-tuning deep networks.
\end{abstract}

\section{Introduction}

In recent years, advances in technology have led to the generation of massive high-dimensional data for computer vision and machine learning. Such data pose challenges, including difficulty in analysis and high computational costs. Fortunately, these high-dimensional data often lie in a union of low-dimensional subspaces. For example, face images captured under varying lighting conditions reside in a nine-dimensional subspace \cite{subspace_lambertian}, and the motion trajectories of rigidly moving objects in a video belong to different subspaces of at most three dimensions \cite{subspace_motion}. To effectively uncover these underlying structures, subspace clustering has emerged as a powerful technique for grouping data points based on their intrinsic subspace membership. Over the past several decades, it has had wide applications in computer vision \citep{sc_cv1}\cite{sc_cv2}\cite{li2015temporal}\cite{multitvc}, image processing \cite{sc_ip1} and systems theory \cite{sc_st1}.


Some classical subspace clustering methods, such as Sparse Subspace Clustering (SSC) \cite{SSC} and Low-Rank Representation (LRR) \cite{LRR}, cluster data drawn from multiple low-dimensional linear or affine subspaces embedded in a high-dimensional space. However, real-world datasets usually contain nonlinear relationships. Some works have tried to nonlinearly map input data into latent space for clustering using kernel tricks \cite{kernel_subspace1} \cite{kernel_subspace2} or autoencoders, but these two-step approaches separate feature learning from clustering objectives. Consequently, the learned representations are not well-adapted to clustering tasks, limiting their effectiveness. 

To bridge this gap, \cite{DSC} introduce the DSC framework where a self-expressive layer resides between the encoder and the decoder, enabling joint optimization of feature learning and clustering. Despite its conceptual simplicity, this method requires processing the entire dataset in a single forward pass, which necessitates downsampling and full-batch training. These constraints limit its applicability to deeper networks and high-resolution data.

Subsequent work \cite{efficientDSC} \cite{deepSC} attempts to circumvent this issue by replacing the self-expressive module with alternative mechanisms, though at the cost of increased implementation complexity. In contrast, our work generalizes DSC to mini-batch training via a memory bank that preserves global feature correlations, enabling scalable training with deeper architectures.

Conventional deep subspace clustering methods adopt shallow networks trained from scratch, while modern pre-trained models like ResNets \cite{resnet} and Vision Transformers (ViTs) \cite{vit} offer richer representations, yet adapting them to subspace clustering is impractical under full-batch training. Although we can build decoders and fine-tune these encoders via mini-batch gradient descent, it yields only marginal improvements, as the features remain unaligned with clustering objectives. Leveraging our mini-batch training strategy, we prove the effectiveness of fine-tuning large-scale pretrained models in subspace clustering. Besides, since the training data for subspace clustering are usually insufficient for training large decoders, we propose a decoder-free architecture that substitutes autoencoding with contrastive learning \cite{simclr}, making it more efficient for fine-tuning deep networks. Experiments on benchmark datasets demonstrate that our method achieves state-of-the-art performance among subspace clustering approaches.

\section{Related work}


\subsection{Mini-batch training}
The adoption of mini-batch optimization in training deep neural networks has been widely studied due to its computational and statistical advantages. Unlike full-batch gradient descent, mini-batch methods partition the dataset into smaller subsets, enabling efficient memory usage and parallelizable computations, which is critical for large-scale vision tasks \cite{efficient_backprop}. Empirical evidence suggests that mini-batch training introduces stochasticity, acting as a regularizer to mitigate overfitting while maintaining convergence stability compared to pure batch gradient descent(BGD). For instance, ResNet \cite{resnet} and other vision architectures leverage mini-batch SGD with momentum to achieve robust feature learning across diverse datasets. Further analysis by Smith et al. \cite{smith2017bayesian} highlights the trade-off between batch size and gradient variance, demonstrating that smaller batches enhance generalization by escaping sharp minima—a key factor in vision model robustness.

\subsection{Self-expressiveness}
Self-expressiveness \cite{SSC} denotes the property that a given data point can be represented as a linear combination of points from the same subspace. Suppose we have data $\mathbf{X} \in \mathcal{R}^{N\times d}$, self-expressiveness can be represented as $\mathbf{X}=\mathbf{C}\mathbf{X}$, where $\mathbf{C}$ is the self-representation coefficient matrix. By applying $l_1$ or $l_2$ regularization on $\mathbf{C}$, $\mathbf{C}$ has a block diagonal structure (after permutation), and each block corresponds to one particular subspace. To avoid each data point being represented by itself, there is another restriction: $diag(\mathbf{C})=\mathbf{0}$. Once $\mathbf{C}$ is obtained, an affinity matrix can be constructed by $|\mathbf{C}|+|\mathbf{C}^T|$ or other heuristics, and used for spectral clustering for final results.

Since self-expressiveness only holds for linear subspaces, \cite{DSC} propose a deep subspace clustering network (DSC) which consists of an autoencoder and a self-expressive layer. The autoencoder helps to learn a latent space through nonlinear mapping, and the self-expressive layer is used to mimic the self-expressiveness property. Under this framework, we can learn representations that are well-suited for subspace clustering and a self-representation coefficient matrix that can be used for spectral clustering.


\subsection{Contrastive learning with a Memory Bank}
In unsupervised learning, contrastive learning often employs a pretext task known as instance discrimination \cite{intsdisc}. For a given image, two data augmentations are applied to generate two variants: $\mathbf{X}$, which serves as the anchor, and $\mathbf{X}^+$, which forms a positive pair with $\mathbf{X}$. Images from different instances, denoted as $\mathbf{X}^-$, are treated as negative samples. The objective of this framework is to minimize the distance between the positive pair representations while maximizing the distance between the anchor and all negative pairs.

The performance of contrastive learning benefits from the number of negative samples. To this end, either a large batch size \cite{simclr} or a memory bank is typically used \cite{intsdisc} \cite{moco2020}. However, using a large batch size often comes with the challenge of excessive memory usage, which is typically unaffordable on most devices. As an alternative, a memory bank that stores the representations of a large number of samples from the dataset offers a more memory-efficient solution. Similarly, traditional DSC methods can be seen as large-batch-based approaches, while our method leverages a memory bank to achieve the same objective more efficiently.

\section{Method}
\begin{figure}[t]
    \centering
    \subfloat[]{
    \label{se_comparison}
        \includegraphics[width=0.3\textwidth]{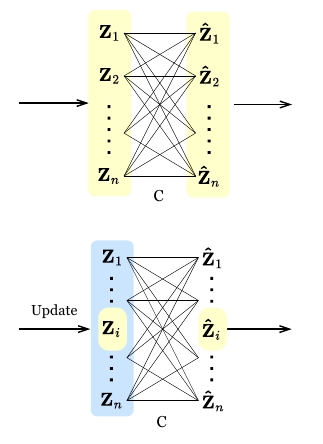}
    }
    \subfloat[]{
    \label{overview_clbdsc}
        \includegraphics[width=0.7\textwidth]{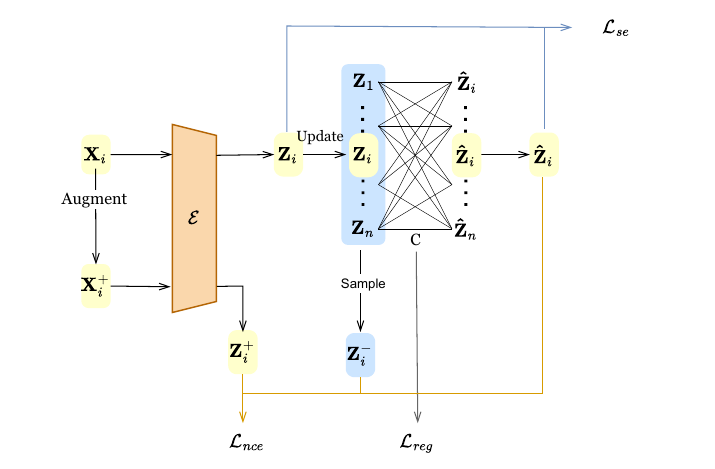}
    }
    \caption{Overview of (a) a comparison between the self-expressive module for DSC (top) and BDSC (bottom). (b) the framework of CLBDSC. Data in the current batch and memory bank are highlighted in yellow and blue, respectively. }
\end{figure}
\subsection{Preliminary}
The deep subspace clustering network (DSC) is composed of an encoder $\mathcal{E}$, a decoder $\mathcal{D}$ and a self-expressive layer parameterized by the matrix $\mathbf{C}$ between them. Given a dataset $\mathbf{X}\in \mathcal{R}^{N \times d}$ containing $N$ samples of $d$ dimensions, in the forward pass, $\mathbf{X}$ is processed by the encoder to acquire its representation $\mathbf{Z}\in \mathcal R^{N\times h}$. The self-expressive layer then reconstructs $\mathbf{Z}$ as $\hat {\mathbf Z} = \mathbf C \mathbf Z$, where $\mathbf {C}\in \mathcal R^{N\times N}, diag(\mathbf C)=\mathbf{0}$ is a matrix capturing linear correlations between samples. Finally, the decoder $\mathcal{D}$ maps $\mathbf{\hat{Z}}$ back to the original image space, generating reconstructed inputs $\mathbf{\hat{X}}$. The DSC framework is optimized to jointly minimize the reconstruction error between $\mathbf{X}$ and $\mathbf{\hat{X}}$, as well as the discrepancy between $\mathbf{Z}$ and $\mathbf{\hat{Z}}$.

A key challenge arises during mini-batch training, where the entire dataset is split into $k$ mini-batches: $\mathbf{X}=[\mathbf{X}_1; \dots; \mathbf{X}_i;\dots;\mathbf{X}_k]$. For a batch $\mathbf{X}_i \in \mathcal{R}^{n \times d}$ where $n$ denotes the batch size, the latent representation $\mathbf{Z}_i\in \mathcal R^{n\times h}$ has reduced dimensionality ($n<N$), making the self-expressive layer inapplicable: its parameter $\mathbf{C}$ inherently requires pairwise relationships across the \textit{entire} dataset rather than individual batches. To address this issue, we introduce a memory bank that persistently maintains latent representations of all training samples. This repository dynamically updates during training, enabling consistent access to global subspace relationships while retaining computational efficiency through batch-wise processing. A comparison of the self-expressive module for DSC and our mini-batch training method (BDSC) is shown in Figure \ref{se_comparison}.



\subsection{Generalized DSC framework}

Our proposed framework extends the architecture of DSC by incorporating an additional memory bank. The training process consists of two stages: pre-training and fine-tuning, where the goal is to first learn meaningful representations and subsequently adapt them for subspace clustering.

\textbf{Pre-training}: In this stage, we disable the self-expressive layer, allowing the autoencoder to independently learn data representations. During the forward process, the encoder $\mathcal{E}$ projects a batch of input samples $\mathbf{X}_i$ into a low-dimensional latent space, producing their latent representation $\mathbf{Z}_i = \mathcal{E}(\mathbf{X}_i)$. The decoder $\mathcal{D}$ then reconstructs this latent variable back to the original space, producing the output $\mathbf{\hat{X}}_i = \mathcal{D}(\mathbf{Z}_i)$. The network is optimized by minimizing the reconstruction loss, specifically the Mean Squared Error (MSE):

\begin{equation}
\label{loss_recon}
    \mathcal{L}_{recon} =  \|\mathbf{X}_i - \mathbf{\hat{X}}_i\|_F^2
\end{equation}

After the pre-training stage, we initialize the feature memory bank by iterating over the dataset and storing the corresponding latent representations $\mathbf{Z} = [\mathbf{Z}^{(0)}_1; \dots; \mathbf{Z}^{(0)}_i;\dots;\mathbf{Z}^{(0)}_k]$. 

\textbf{Fine-tuning}: In the fine-tuning stage, the self-expressive layer is activated to refine the latent representations by enforcing structural constraints. For the \(i\)-th batch in epoch \(t\), the encoder generates latent features \(\mathbf{Z}_i^{(t)} = \mathcal{E}(\mathbf{X}_i^{(t)})\) which is used to update the memory bank as \(\mathbf{Z} = [\mathbf{Z}_1^{(t)}; \ldots; \mathbf{Z}_i^{(t)}; \ldots; \mathbf{Z}_k^{(t-1)}]\). The self-expressive layer reconstructs the current representation as \(\mathbf{\hat{Z}}_i^{(t)} = (\mathbf{C}\mathbf{Z})_i\), using both current and historical representations. Finally, the decoder recovers samples from these latent representations as \(\mathbf{\hat{X}}_i = \mathcal{D}(\mathbf{\hat{Z}}_i^{(t)})\), enabling joint optimization of both reconstruction accuracy and subspace clustering consistency via gradient backpropagation. The loss function for optimizing the self-expressive layer is:

\begin{equation}
    \mathcal{L}_{se} =  \|\mathbf{Z}_i - \mathbf{\hat{Z}}_i\|_F^2 \quad \textit{s.t. }diag(\mathbf{C})=\mathbf{0}
\end{equation}

For regularization, we adopt $l_2$ norm on $\mathbf{C}$, i.e., $\mathcal{L}_{reg}=\|\mathbf{C}\|_2$ . The overall loss function is a weighted sum of the reconstruction loss, the self-expressive loss, and the regularization term:
\begin{equation}
\label{loss_ae}
    \mathcal{L} = \mathcal{L}_{recon} +\alpha\mathcal{L}_{se}+\beta\mathcal{L}_{reg}
\end{equation}
where $\alpha, \beta$ are balancing coefficients. After back propagation, $\mathbf{Z}$ is detached from the computational graph. Notably, when the batch size is set to $n=N$, BDSC reduces to DSC, as the entire memory bank is updated at each iteration.

\subsection{Consistency for memory bank}

For learning a good self-expressive coefficient matrix, we need to ensure that features come from similar encoders. This issue does not arise in full-batch methods, but in mini-batch training, features in the memory bank are about the encoders at multiple different steps over the past epoch, and are less consistent.

When the number of splits $k$ is large, features from the first batch and the last batch are likely to be encoded by significantly different encoders. This can significantly disrupt the learning of self-expressive coefficients. In such cases, the features from the first batch may become "outdated" and unsuitable for representing newer features. Our goal is to maintain consistency between features from different batches. Since the size of subspace clustering datasets is not large (usually < 10k), for simplicity, we do not adopt a momentum encoder \cite{moco2020} and instead mitigate the issue by reducing the encoder’s learning rate, thereby slowing down the updates to its parameters. In practice, the relationship between the learning rate $lr$ and the number of splits $k$ is given by $lr \propto  \frac{1}{k}$. In other words, as the number of splits increases, we reduce the learning rate to ensure consistent feature encoding.

\subsection{Decoder-free framework}
Many large-scale pre-trained models are encoder-only. However, subspace clustering datasets are typically small, making them inadequate for training large decoders. This raises the question whether DSC can be generalized to a decoder-free framework. In DSC, the decoder plays a critical role in learning the latent space, but recent self-supervised methods, such as contrastive learning, offer an alternative approach. In this section, we propose a contrastive learning-based DSC network that can also be optimized using mini-batches. Within this framework, our memory bank can also serve as a source for sampling negative examples.

Unlike previous contrastive learning works \cite{simclr} \cite{byol}, we retain the original images $\mathbf{X}_i$ and apply a single augmentation to obtain ${\mathbf{X}}_i^+$. In the forward pass, we encode both $\mathbf{X}_i$ and ${\mathbf{X}}_i^+$ to generate their respective representations $\mathbf{Z}_i$ and $\mathbf{Z}_i^+$. The representation $\mathbf{Z}_i$ is then passed through the self-expressive module. To compute the contrastive loss, we treat $\mathbf{\hat{Z}}_i$ as the anchor, $\mathbf{Z}_i^+$ as the positive pair, and $\mathbf{Z}^-_i$ from the memory bank as negative pairs. The InfoNCE loss \cite{infonce} is adopted:

\begin{equation}
    \mathcal{L}_{nce} = - \log \frac{\exp(\hat{\mathbf{z}}\cdot \mathbf{z}^+/\tau)}
    {\sum_{k=0}^{N-n}\exp(\mathbf{\hat{z}}\cdot \mathbf{z}_k^-/\tau)}
\end{equation}

where $\tau$ is the temperature, and $\mathbf{\hat{z}}$, $\mathbf{z}^+$ , $\mathbf{z}_k^-$ denote representations from $\mathbf{\hat{Z}}_i$, $\mathbf{Z}_i^+$, $\mathbf{Z}^-_i$ (all normalized to unit length). As $N$ is not excessively large, all samples from the memory bank, except the current batch, are treated as negative samples. We take $\mathbf{\hat{z}}$ instead of $\mathbf{z}$ as the anchor to prevent the decoupling of representation learning from subspace clustering. To ensure that each representation in the memory bank corresponds to a deterministic input, we update the memory bank with $\mathbf{z}$ rather than $\mathbf{z}^+$. 

The final objective function shares the structure of equation (\ref{loss_ae}), but $\mathcal{L}_{recon}$ is replaced with $\mathcal{L}_{nce}$. We denote this contrastive learning-based approach as CLBDSC. Its framework is presented in Figure \ref{overview_clbdsc}.

\section{Experiment}
\label{sec_experiment}
\subsection{Experimental setups}
\textbf{Datasets}: Since classical subspace clustering methods and deep subspace clustering methods have quadratic training time and space complexities in terms of the number of samples, we focus on using deep architectures and high-resolution for small or medium-scale datasets (sample < 10k). Following previous works, four subspace clustering datasets are used for evaluating our method, including ORL, COIL-20, COIL-100 and Extended Yale-B.

ORL \cite{ORL} has 400 face images of 40 people in $98 \times112$ grayscale with different expression details such as wearing glasses or not, smiling or not, eyes open or closed. For each subject, the images were taken under varying lighting conditions.

COIL-100 and COIL-20 \cite{COIL20} datasets consist of images of 100 and 20 objects, respectively, placed on a motorized turntable. For each object, 72 images are taken at poses intervals of 5 degrees that cover a 360-degree range. The COIL-100 dataset contains RGB images, whereas the COIL-20 dataset consists of grayscale images.

The Extended Yale-B dataset \cite{yaleb} contains 2432 facial images of 38 individuals from 9 poses and under 64 illumination settings.

\textbf{Evaluating metrics}: We choose two popular metrics, i.e. clustering accuracy rate (ACC), normalized mutual information (NMI), to evaluate the performance of our models. Both metrics take values in [0,1], with higher values indicating better performance.

ACC is defined as:
\begin{equation}
ACC=\frac{\sum_{i=1}^{N}\mathbbm{1}(p_i,r_i)}{N} 
\end{equation}
where $p_i$ is the predicted label for the $i_{th}$ sample, and $r_i$ is the real label. $\mathbbm{1}(\cdot,\cdot)$ is the indicator function that satisfies $\mathbbm{1}(p_i,r_i)=1$ if $p_i=r_i$ and $\mathbbm{1}(p_i,r_i)=0$ otherwise.

NMI is defined as follows:
\begin{equation}
NMI=\frac{2\mathbf{I}(p,r)}{\mathbf{H}(p)+\mathbf{H}(r)} 
\end{equation}
where $\mathbf{I}(\cdot,\cdot)$ is the mutual information. $\mathbf{H}$ represents the entropy.

Our proposed method is implemented using PyTorch 2.0.1. We adopt Adam \cite{adam} as the optimizer with default momentum parameters. For CLBDSC, we pre-train the self-expressive layer with frozen encoders before full fine-tuning. All the experiments are conducted on one NVIDIA RTX 3080Ti.


\subsection{Approximating Full-Batch Training with mini-batch Strategies}
\label{sec_approx}

To assess whether our mini-batch training strategy achieves performance comparable to full-batch methods, we conduct experiments using architectures and hyper-parameters identical to those of DSC \cite{DSC} and convert all datasets to grayscale. Specifically, images from ORL, COIL20/100 are resized to \(32\times32\), and Yale-B to \(42\times42\). The batch size is set to 32 for ORL, COIL20, 128 for Yale-B, and 256 for COIL100 to ensure computational efficiency. For improving the consistency in the feature memory bank, we set a small learning rate of $0.0001$.  

As shown in Table~\ref{approx}, our method (BDSC) achieves performance comparable to that of full-batch DSC, sometimes surpassing it, demonstrating that our mini-batch training is a good alternative to traditional full-batch methods. Results for ORL, COIL20, and Yale-B are based on both the publicly available PyTorch implementation\footnote{https://github.com/XifengGuo/DSC-Net.git} of DSC and our reimplementation. While for COIL100, we report results based on the original TensorFlow implementation.

\begin{table}[h]
    \centering
    \caption{Comparison with full-batch strategy.}
    \begin{tabular}{ccccc}
    \toprule
           &Metric& DSC(reported) & DSC(our implement) & BDSC \\ \midrule
\multirow{2}{*}{ORL}&ACC&  0.855&0.863& 0.888\\ 
                    &NMI&  0.920&0.918& 0.950\\ \midrule
\multirow{2}{*}{COIL20}&ACC&  0.910&0.916& 0.919\\
                       &NMI&  0.959&0.961& 0.972\\ \midrule
\multirow{2}{*}{COIL100}&ACC&  0.690&0.632& 0.666\\ 
                        &NMI&  -&0.887& 0.896\\ \midrule
\multirow{2}{*}{Extended Yale-B}&ACC&  0.973&0.973& 0.959\\ 
                        &NMI&  0.963&0.963& 0.946\\ 
    \bottomrule
    \end{tabular}
    \label{approx}
\end{table}


\subsection{Experiments with high-resolution images and deeper networks}
\label{sec_deeper}

Our mini-batch training strategy significantly reduces the memory cost and thus facilitates the deployment of deeper architectures and higher-resolution inputs. For evaluating whether subspace clustering performance benefits from deeper networks, in this section, we conduct experiments on ORL and COIL100 datasets. The architectures used in this section are detailed in the Appendix.

\begin{table}[h]
    \caption{Augmentations applied in CLBDSC for each dataset}
    \label{aug}
    \centering
    \begin{tabular}{c|cc}\toprule
    Dataset     & COIL100&  ORL\\ \midrule
    \multirow{4}{*}{Augment}     
    & Crop and Resize & Crop and Resize\\
    &Color Jitter  & Gray Scale\\   
    &Random Grayscale         &Horizontal Flip\\
    &Horizontal Flip                    \\ \bottomrule
    \end{tabular}
\end{table}
\paragraph{ORL} For BDSC, we employ a 3-layer encoder with batch normalization \citet{BN} applied after each convolutional layer, followed by GELU \citet{GELU} activation. The decoder mirrors the encoder but omits batch normalization. We flatten the features extracted by the encoder before the self-expressive module. Input images of size \(32 \times 32\) are encoded into 256-dimensional latent features. We set the learning rate to $1\times10^{-3}$, $\alpha$ to 50 and $\beta$ to 1. The autoencoder is first pre-trained without the self-expressive module for 100 epochs, and then fine-tuned for 1000 epochs.

For CLBDSC, we utilize \(128 \times 128\) resolution inputs and apply the augmentations detailed in Table~\ref{aug} to generate positive pairs, including random cropping (with resizing), color distortion, horizontal flipping and random grayscale \citet{color_resize} \citet{other_aug} \citet{color_resize}. We adopt 6-layer encoder and set the learning rate to $5\times10^{-5}$, keeping other settings the same with BDSC.

\paragraph{COIL100} For BDSC, we design an autoencoder containing an ImageNet-1k pre-trained ResNet-18 as encoder and a lightweight decoder containing only 6M parameters, which is smaller than the encoder (11M). This design prioritizes efficient representation learning over reconstruction quality, as better reconstruction does not necessarily indicate better representation learning. Batch normalization and ReLU activation follow each deconvolutional layer.  

For CLBDSC, we adopt the same ResNet-18 encoder with \(128 \times 128\) inputs and augmentations from Table~\ref{aug}. For both methods, given that the encoder is pre-trained, we directly fine-tune the model for 80 epochs. A learning of $5\times10^{-5}$ is used, with $\alpha$, $\beta$ set to 1.

\begin{figure}[h] 
    \centering
    \begin{subfigure}[b]{0.45\textwidth} 
        \centering
        \includegraphics[width=1\textwidth]{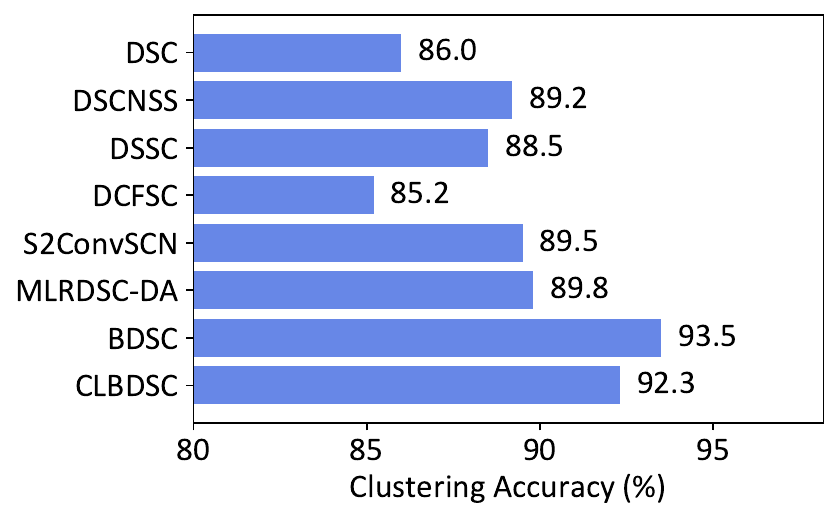} 
        \caption{ORL} 
    \end{subfigure}
    \hfill 
    \begin{subfigure}[b]{0.45\textwidth} 
        \centering 
        \includegraphics[width=1\textwidth]{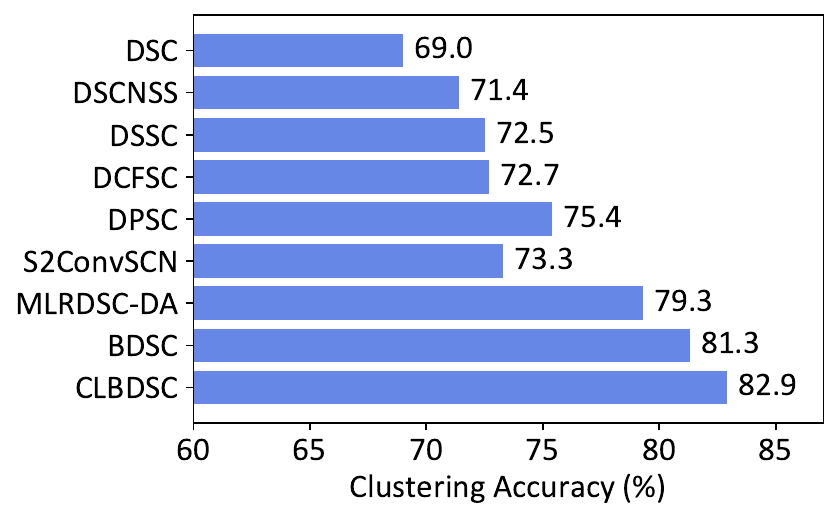} 
        \caption{COIL-100} 
    \end{subfigure}
    \caption{Comparison with SOTA subspace clustering methods on COIL100 and ORL}
    \label{Com_SOTA_orl_coil100}
\end{figure}

As shown in Figure \ref{Com_SOTA_orl_coil100}, our methods outperform previous state-of-the-art subspace clustering approaches, including DSCNSS \cite{DSCNSS_2023}, DSSC \cite{DSSC_2021}, DCFSC \cite{DCFSC_2019}, DPSC \cite{DPSC_2019}, S\(^2\)ConvSCN \cite{cnnSC}, and MLRDSC-DA \cite{augSC}. These results underscore the effectiveness of our mini-batch training strategy and decoder-free contrastive learning framework in deeper networks.  

\begin{figure}[h] 
    \centering
    \begin{subfigure}[b]{0.32\textwidth} 
        \centering 
        \includegraphics[width=1\textwidth]{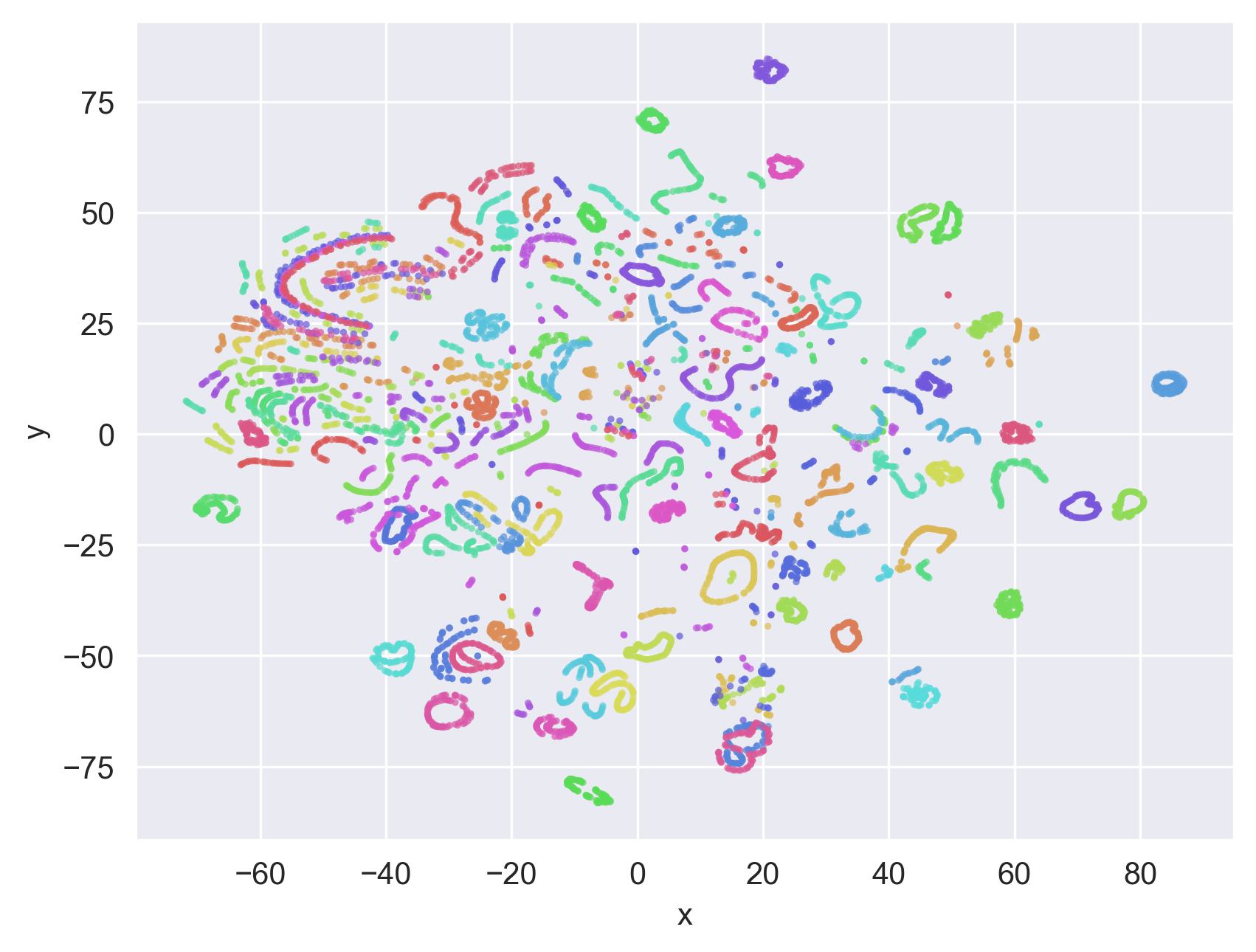} 
        \caption{DSC} 
    \end{subfigure}
    \hfill 
        \hfill 
    \begin{subfigure}[b]{0.32\textwidth} 
        \centering 
        \includegraphics[width=1\textwidth]{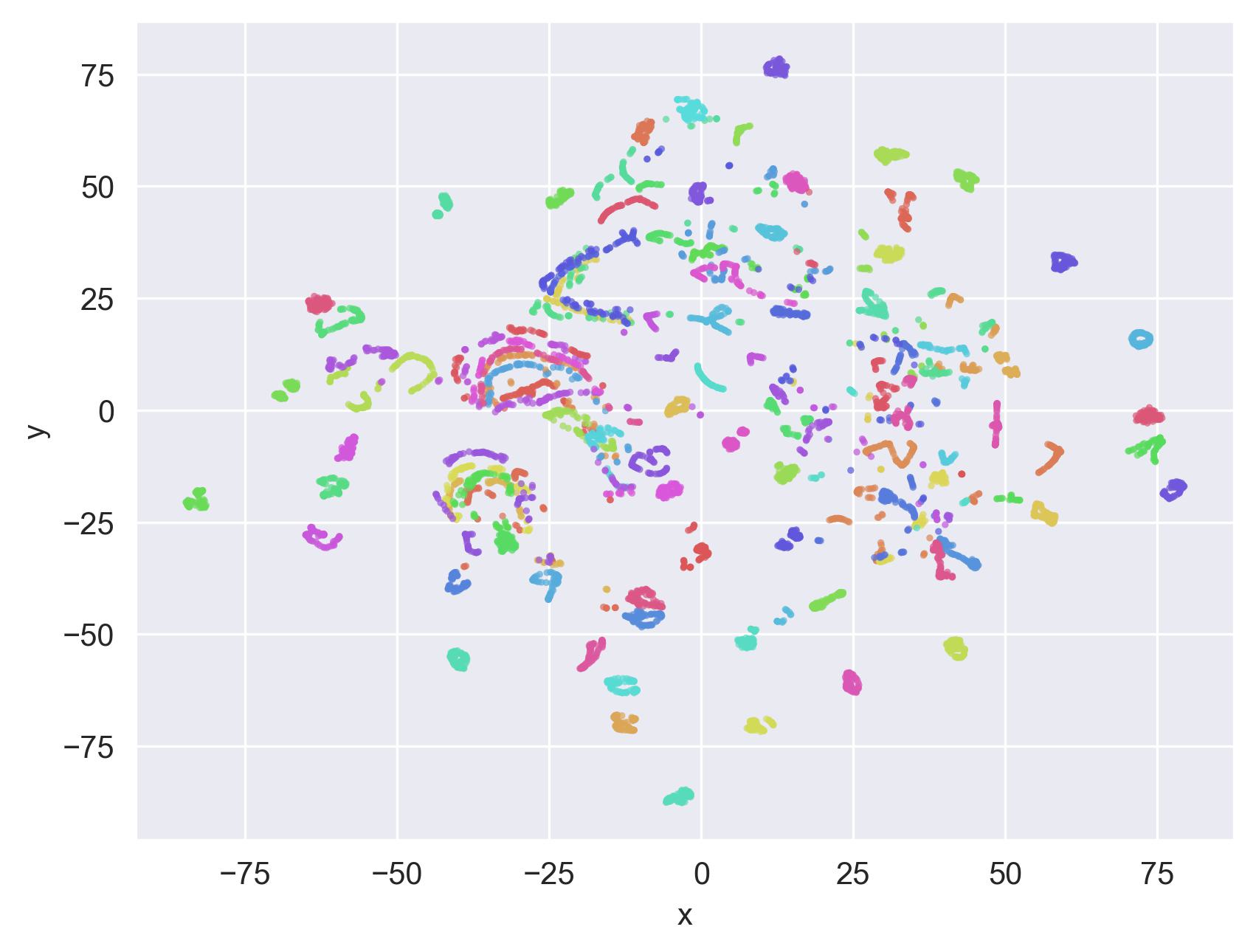} 
        \caption{BDSC} 
    \end{subfigure}
    \hfill 
    \begin{subfigure}[b]{0.32\textwidth} 
        \centering 
        \includegraphics[width=1\textwidth]{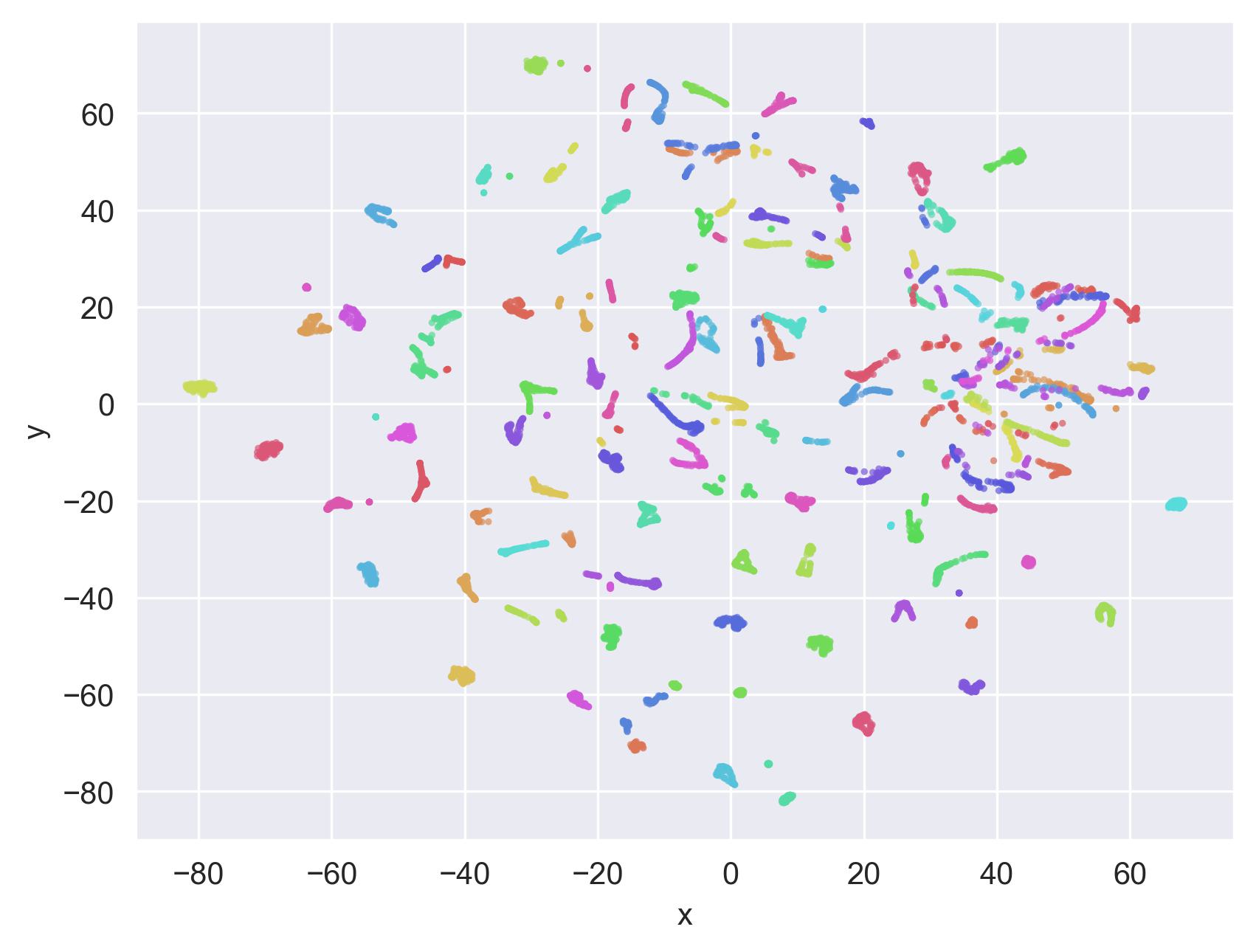} 
        \caption{CLBDSC} 
    \end{subfigure}
    \caption{Visualization of the embedded representations with t-SNE on COIL100. Samples from different classes are marked in different colors.}
    \label{tsne}
\end{figure}

\paragraph{Visualization of Embedded Representation.} Figure \ref{tsne} presents the t-SNE visualization of the learned embedded representation on COIL100. We compare DSC \cite{DSC} with our two models based on ResNet18. Representations learned by our deeper models are apparently distributed in tighter clusters, especially for CLBDSC, which demonstrates the ability to learn more discriminative representations.

\subsection{Ablation study}
\begin{figure}[h] 
    \centering
    \begin{subfigure}[b]{0.45\textwidth}
        \centering
        \includegraphics[width=1\textwidth]{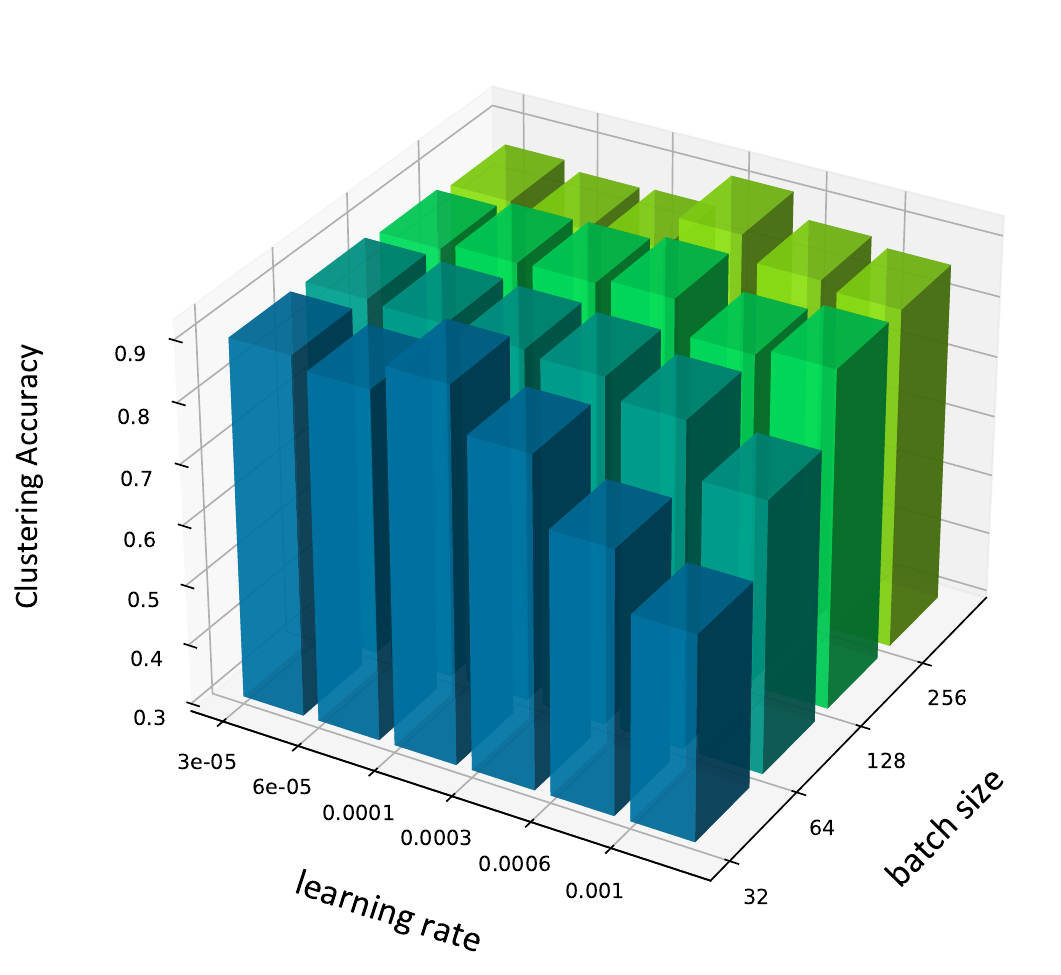}
        \caption{Clustering Accuracy}
    \end{subfigure}
    \hfill
    \begin{subfigure}[b]{0.45\textwidth} 
        \centering 
        \includegraphics[width=1\textwidth]{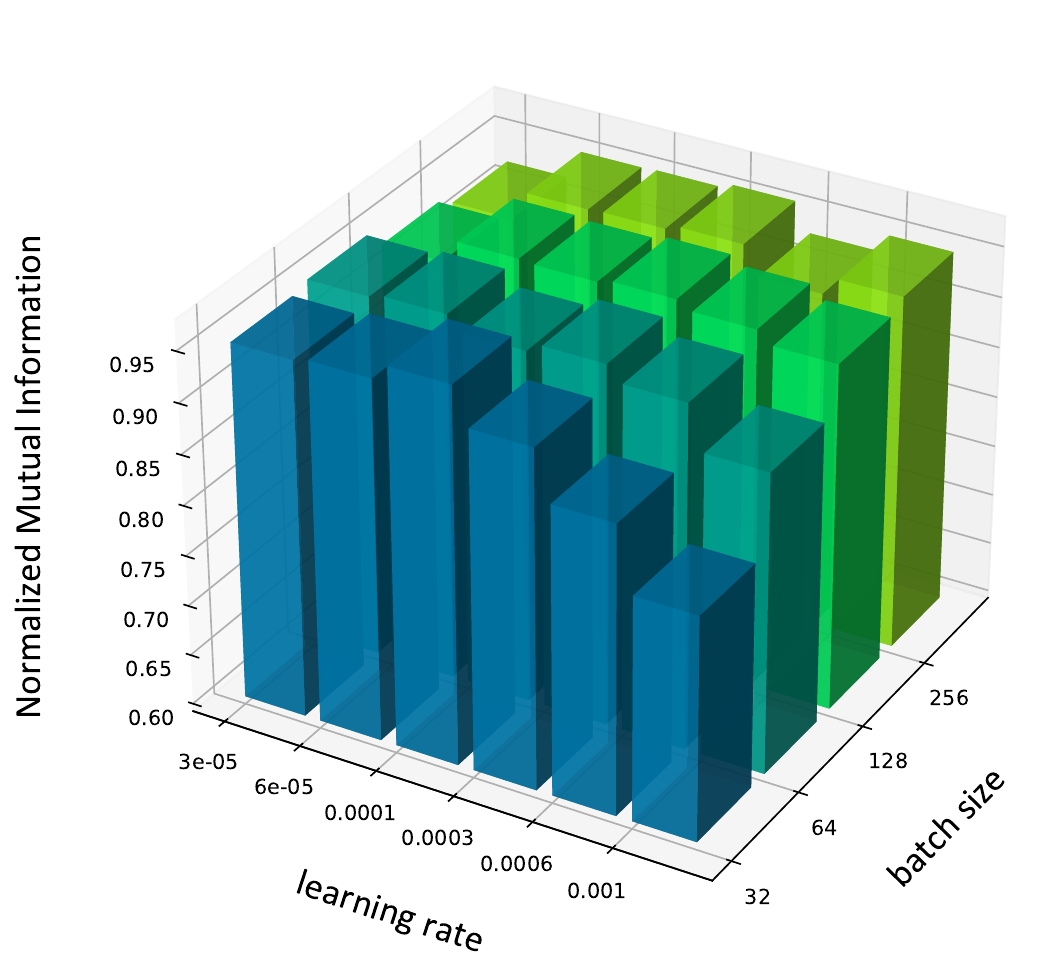} 
        \caption{Normalized Mutual Information} 
    \end{subfigure}
    \caption{Influence of batch size and learning rate.}
    \label{lr_bs}
\end{figure}

\paragraph{Influence of learning rate and batch size.} To validate our analysis on memory bank consistency, we vary the batch size \(\{32, 64, 128, 256\}\) and learning rate \(\{3 \times 10^{-5}, 6 \times 10^{-5}, 10^{-4}, 3 \times 10^{-4}, 6 \times 10^{-4}, 10^{-3}\}\) on COIL20, while keeping other settings identical to Section \ref{sec_approx}. Since we have 1,440 samples in total, the maximum number of splits per epoch reaches 45 when using the smallest batch size (32). Figure \ref{lr_bs} shows the results, revealing two key trends: (1) smaller batch sizes require lower learning rates to improve memory bank consistency, and (2) results with larger batch sizes exhibit less sensitivity to the choice of learning rate due to fewer dataset splits.

\begin{figure}[h] 
    \centering
    \begin{subfigure}[b]{0.45\textwidth} 
        \centering
        \includegraphics[width=1\textwidth]{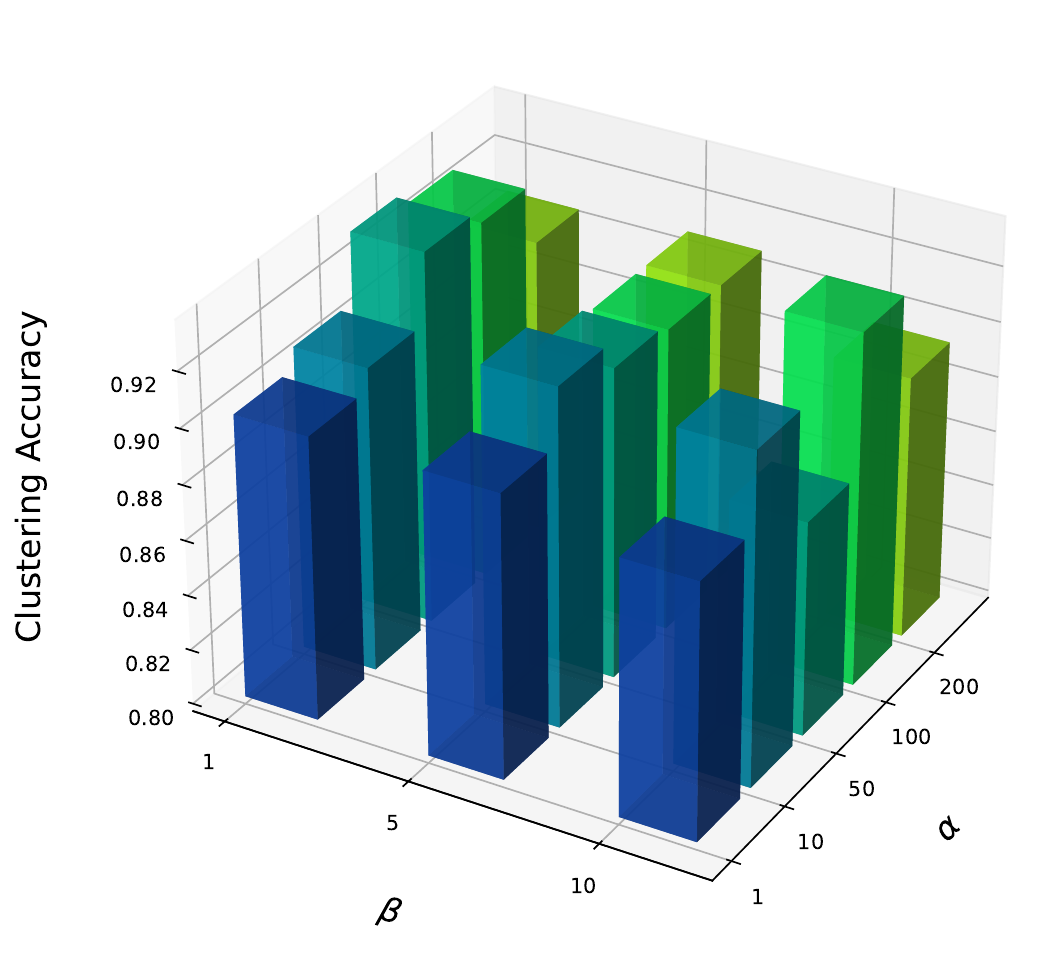} 
        \caption{Clustering Accuracy} 
        \label{lr}
    \end{subfigure}
    \hfill 
    \begin{subfigure}[b]{0.45\textwidth} 
        \centering 
        \includegraphics[width=1\textwidth]{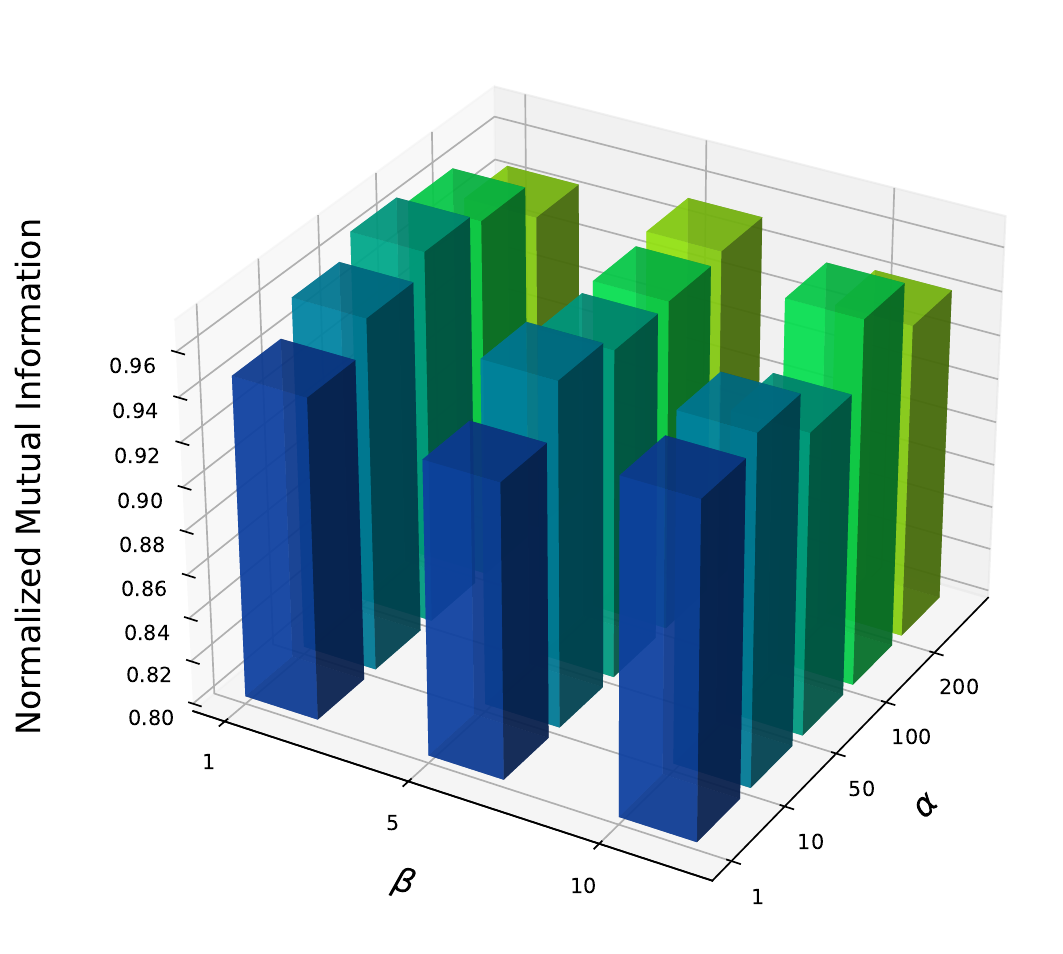} 
        \caption{Normalized Mutual Information} 
        \label{bs}
    \end{subfigure}
    \caption{Influence of balancing coefficients}
    \label{coefficients}
\end{figure}

\paragraph{Influence of balancing coefficients.} We analyze the impact of loss-balancing coefficients \(\alpha\) and \(\beta\) by testing \(\alpha \in \{1, 10, 50, 100, 200\}\) and \(\beta \in \{1, 5, 10\}\) on ORL for computational efficiency. As shown in Figure \ref{coefficients}, a small \(\beta\) and a moderate \(\alpha\) (around 50) enhance the performance. However, these parameters have a relatively minor effect on final results compared to the learning rate.



\paragraph{Influence of multi-task training} Our framework jointly optimizes representation learning and clustering, making it a form of multi-task training. To assess its impact, we compare our methods (BDSC, CLBDSC) with two-step training approaches: (1) clustering (typically via SSC) on features extracted from pre-trained or fine-tuned autoencoders (AE+SSC), and (2) clustering on features learned by contrastive learning (CL+SSC). 

As shown in Table~\ref{ablation}, joint training significantly outperforms two-stage approaches. On COIL100, fine-tuning an autoencoder yields only a 3.7\% accuracy improvement over pre-trained features, whereas contrastive learning (CL+SSC) achieves a 9\% increase. However, integrating clustering objectives directly into training, as in BDSC and CLBDSC, results in accuracies exceeding 80\%. A similar trend is observed on ORL, where multi-task optimization consistently outperforms two-step training, highlighting the importance of unifying representation learning and subspace clustering.



\begin{table}[h]
    \centering
    \caption{Comparison with two-step training.}
    \begin{tabular}{cccc} \toprule
    Dataset             &Method    & ACC&  NMI \\ \midrule
    \multirow{5}{*}{COIL100}  
    & ResNet18 (pre-trained)+SSC            & 0.661 &0.906\\
    & ResNet18 (fine-tuned with AE)+SSC     & 0.698 &0.936 \\ 
    &BDSC                                   & 0.813 &0.963 \\ 
    &CL+SSC                                 & 0.752 &0.948 \\ 
    &CLBDSC                                 & 0.829 &0.964 \\\midrule
    \multirow{4}{*}{ORL}
    &AE+SSC              & 0.853 &0.909 \\
    &BDSC                & 0.935 &0.970 \\
    &CL+SSC              & 0.905 &0.948 \\ 
    &CLBDSC              & 0.923 &0.955 \\
    \bottomrule
    \end{tabular}
    \label{ablation}
\end{table}

\section{Conclusion}
In this paper, we propose a mini-batch training strategy for deep subspace clustering networks. Compared with traditional full-batch methods, our memory bank-based approach achieves competitive performance with lower memory cost. Based on it, we design a decoder-free framework which replaces autoencoding with contrastive learning, containing less parameters while achieving comparable results. Extensive experiments demonstrate the feasibility and effectiveness of fine-tuning large-scale pre-trained models for subspace clustering. For now, our method cannot deal with large scale datasets, mainly due to the limitation of self-expressive subspace clustering. In future work, we aim to address this problem to achieve broader applicability.

\section{Broader Impact}
Our mini-batch DSC framework enables efficient subspace clustering of high-dimensional visual data like medical images and satellite imagery. By eliminating the need for labeled data, the method could benefit domains where annotation is expensive or impractical. Potential considerations include: privacy implications when processing sensitive unlabeled datasets (e.g., facial images), and biases in cluster structures that may emerge from pre-trained encoders. Besides our decoder-free design reduces energy consumption compared to traditional autoencoder approaches, which is more environmentally friendly.

\bibliographystyle{abbrvnat}
\bibliography{sec/reference}

\begin{thebibliography}{38}
\providecommand{\natexlab}[1]{#1}
\providecommand{\url}[1]{\texttt{#1}}
\expandafter\ifx\csname urlstyle\endcsname\relax
  \providecommand{\doi}[1]{doi: #1}\else
  \providecommand{\doi}{doi: \begingroup \urlstyle{rm}\Url}\fi

\bibitem[Abavisani et~al.(2020)Abavisani, Naghizadeh, Metaxas, and Patel]{augSC}
M.~Abavisani, A.~Naghizadeh, D.~Metaxas, and V.~Patel.
\newblock Deep subspace clustering with data augmentation.
\newblock \emph{Advances in Neural Information Processing Systems}, 33:\penalty0 10360--10370, 2020.

\bibitem[Baek et~al.(2021)Baek, Yoon, Song, and Yoon]{DSSC_2021}
S.~Baek, G.~Yoon, J.~Song, and S.~M. Yoon.
\newblock Deep self-representative subspace clustering network.
\newblock \emph{Pattern Recognition}, 118:\penalty0 108041, 2021.

\bibitem[Basri and Jacobs(2003)]{subspace_lambertian}
R.~Basri and D.~W. Jacobs.
\newblock Lambertian reflectance and linear subspaces.
\newblock \emph{IEEE transactions on pattern analysis and machine intelligence}, 25\penalty0 (2):\penalty0 218--233, 2003.

\bibitem[Cai et~al.(2022)Cai, Fan, Guo, Wang, Zhang, and Zhang]{efficientDSC}
J.~Cai, J.~Fan, W.~Guo, S.~Wang, Y.~Zhang, and Z.~Zhang.
\newblock Efficient deep embedded subspace clustering.
\newblock In \emph{Proceedings of the IEEE/CVF conference on computer vision and pattern recognition}, pages 1--10, 2022.

\bibitem[Chen et~al.(2023)Chen, Lu, Wei, and Geng]{DSCNSS_2023}
C.~Chen, H.~Lu, H.~Wei, and X.~Geng.
\newblock Deep subspace image clustering network with self-expression and self-supervision.
\newblock \emph{Applied Intelligence}, 53\penalty0 (4):\penalty0 4859--4873, 2023.

\bibitem[Chen et~al.(2020)Chen, Kornblith, Norouzi, and Hinton]{simclr}
T.~Chen, S.~Kornblith, M.~Norouzi, and G.~Hinton.
\newblock A simple framework for contrastive learning of visual representations.
\newblock In \emph{International conference on machine learning}, pages 1597--1607. PmLR, 2020.

\bibitem[Dosovitskiy et~al.(2020)Dosovitskiy, Beyer, Kolesnikov, Weissenborn, Zhai, Unterthiner, Dehghani, Minderer, Heigold, Gelly, et~al.]{vit}
A.~Dosovitskiy, L.~Beyer, A.~Kolesnikov, D.~Weissenborn, X.~Zhai, T.~Unterthiner, M.~Dehghani, M.~Minderer, G.~Heigold, S.~Gelly, et~al.
\newblock An image is worth 16x16 words: Transformers for image recognition at scale.
\newblock \emph{arXiv preprint arXiv:2010.11929}, 2020.

\bibitem[Grill et~al.(2020)Grill, Strub, Altch{\'e}, Tallec, Richemond, Buchatskaya, Doersch, Avila~Pires, Guo, Gheshlaghi~Azar, et~al.]{byol}
J.-B. Grill, F.~Strub, F.~Altch{\'e}, C.~Tallec, P.~Richemond, E.~Buchatskaya, C.~Doersch, B.~Avila~Pires, Z.~Guo, M.~Gheshlaghi~Azar, et~al.
\newblock Bootstrap your own latent-a new approach to self-supervised learning.
\newblock \emph{Advances in neural information processing systems}, 33:\penalty0 21271--21284, 2020.

\bibitem[He et~al.(2016)He, Zhang, Ren, and Sun]{resnet}
K.~He, X.~Zhang, S.~Ren, and J.~Sun.
\newblock Deep residual learning for image recognition.
\newblock In \emph{Proceedings of the IEEE conference on computer vision and pattern recognition}, pages 770--778, 2016.

\bibitem[He et~al.(2020)He, Fan, Wu, Xie, and Girshick]{moco2020}
K.~He, H.~Fan, Y.~Wu, S.~Xie, and R.~Girshick.
\newblock Momentum contrast for unsupervised visual representation learning.
\newblock In \emph{Proceedings of the IEEE/CVF conference on computer vision and pattern recognition}, pages 9729--9738, 2020.

\bibitem[Hendrycks and Gimpel(2016)]{GELU}
D.~Hendrycks and K.~Gimpel.
\newblock Gaussian error linear units (gelus).
\newblock \emph{arXiv preprint arXiv:1606.08415}, 2016.

\bibitem[Hong et~al.(2006)Hong, Wright, Huang, and Ma]{sc_ip1}
W.~Hong, J.~Wright, K.~Huang, and Y.~Ma.
\newblock Multiscale hybrid linear models for lossy image representation.
\newblock \emph{IEEE Transactions on Image Processing}, 15\penalty0 (12):\penalty0 3655--3671, 2006.

\bibitem[Ioffe and Szegedy(2015)]{BN}
S.~Ioffe and C.~Szegedy.
\newblock Batch normalization: Accelerating deep network training by reducing internal covariate shift.
\newblock In \emph{International conference on machine learning}, pages 448--456. pmlr, 2015.

\bibitem[Ji et~al.(2017)Ji, Zhang, Li, Salzmann, and Reid]{DSC}
P.~Ji, T.~Zhang, H.~Li, M.~Salzmann, and I.~Reid.
\newblock Deep subspace clustering networks.
\newblock In I.~Guyon, U.~V. Luxburg, S.~Bengio, H.~Wallach, R.~Fergus, S.~Vishwanathan, and R.~Garnett, editors, \emph{Advances in Neural Information Processing Systems}, volume~30. Curran Associates, Inc., 2017.

\bibitem[Kingma and Ba(2014)]{adam}
D.~P. Kingma and J.~Ba.
\newblock Adam: A method for stochastic optimization.
\newblock \emph{arXiv preprint arXiv:1412.6980}, 2014.

\bibitem[Krizhevsky et~al.(2012)Krizhevsky, Sutskever, and Hinton]{color_resize}
A.~Krizhevsky, I.~Sutskever, and G.~E. Hinton.
\newblock Imagenet classification with deep convolutional neural networks.
\newblock \emph{Advances in neural information processing systems}, 25, 2012.

\bibitem[LeCun et~al.(2002)LeCun, Bottou, Orr, and M{\"u}ller]{efficient_backprop}
Y.~LeCun, L.~Bottou, G.~B. Orr, and K.-R. M{\"u}ller.
\newblock Efficient backprop.
\newblock In \emph{Neural networks: Tricks of the trade}, pages 9--50. Springer, 2002.

\bibitem[Lee et~al.(2005)Lee, Ho, and Kriegman]{yaleb}
K.-C. Lee, J.~Ho, and D.~J. Kriegman.
\newblock Acquiring linear subspaces for face recognition under variable lighting.
\newblock \emph{IEEE Transactions on pattern analysis and machine intelligence}, 27\penalty0 (5):\penalty0 684--698, 2005.

\bibitem[Li et~al.(2015)Li, Li, and Fu]{li2015temporal}
S.~Li, K.~Li, and Y.~Fu.
\newblock Temporal subspace clustering for human motion segmentation.
\newblock In \emph{Proceedings of the IEEE international conference on computer vision}, pages 4453--4461, 2015.

\bibitem[Li et~al.(2024)Li, Wang, Li, Yuan, and Wang]{deepSC}
Y.~Li, S.~Wang, C.~Li, Y.~Yuan, and G.~Wang.
\newblock Towards very deep representation learning for subspace clustering.
\newblock \emph{IEEE Transactions on Knowledge and Data Engineering}, 36\penalty0 (7):\penalty0 3568--3579, 2024.

\bibitem[Liu et~al.(2010)Liu, Lin, and Yu]{LRR}
G.~Liu, Z.~Lin, and Y.~Yu.
\newblock Robust subspace segmentation by low-rank representation.
\newblock In \emph{Proceedings of the 27th international conference on machine learning (ICML-10)}, pages 663--670, 2010.

\bibitem[Nene et~al.(1996)Nene, Nayar, Murase, et~al.]{COIL20}
S.~A. Nene, S.~K. Nayar, H.~Murase, et~al.
\newblock Columbia object image library (coil-20).
\newblock 1996.

\bibitem[Oord et~al.(2018)Oord, Li, and Vinyals]{infonce}
A.~v.~d. Oord, Y.~Li, and O.~Vinyals.
\newblock Representation learning with contrastive predictive coding.
\newblock \emph{arXiv preprint arXiv:1807.03748}, 2018.

\bibitem[Patel and Vidal(2014)]{kernel_subspace1}
V.~M. Patel and R.~Vidal.
\newblock Kernel sparse subspace clustering.
\newblock In \emph{2014 ieee international conference on image processing (icip)}, pages 2849--2853. IEEE, 2014.

\bibitem[Patel et~al.(2015)Patel, Van~Nguyen, and Vidal]{kernel_subspace2}
V.~M. Patel, H.~Van~Nguyen, and R.~Vidal.
\newblock Latent space sparse and low-rank subspace clustering.
\newblock \emph{IEEE Journal of Selected Topics in Signal Processing}, 9\penalty0 (4):\penalty0 691--701, 2015.

\bibitem[Samaria and Harter(1994)]{ORL}
F.~S. Samaria and A.~C. Harter.
\newblock Parameterisation of a stochastic model for human face identification.
\newblock In \emph{Proceedings of 1994 IEEE workshop on applications of computer vision}, pages 138--142. IEEE, 1994.

\bibitem[Seo et~al.(2019)Seo, Koo, and Jeon]{DCFSC_2019}
J.~Seo, J.~Koo, and T.~Jeon.
\newblock Deep closed-form subspace clustering.
\newblock In \emph{Proceedings of the IEEE/CVF International Conference on Computer Vision Workshops}, pages 0--0, 2019.

\bibitem[Smith and Le(2017)]{smith2017bayesian}
S.~L. Smith and Q.~V. Le.
\newblock A bayesian perspective on generalization and stochastic gradient descent.
\newblock \emph{arXiv preprint arXiv:1710.06451}, 2017.

\bibitem[Szegedy et~al.(2015)Szegedy, Liu, Jia, Sermanet, Reed, Anguelov, Erhan, Vanhoucke, and Rabinovich]{other_aug}
C.~Szegedy, W.~Liu, Y.~Jia, P.~Sermanet, S.~Reed, D.~Anguelov, D.~Erhan, V.~Vanhoucke, and A.~Rabinovich.
\newblock Going deeper with convolutions.
\newblock In \emph{Proceedings of the IEEE conference on computer vision and pattern recognition}, pages 1--9, 2015.

\bibitem[Tomasi and Kanade(1992)]{subspace_motion}
C.~Tomasi and T.~Kanade.
\newblock Shape and motion from image streams under orthography: a factorization method.
\newblock \emph{International journal of computer vision}, 9:\penalty0 137--154, 1992.

\bibitem[Vidal(2009)]{SSC}
E.~E.~R. Vidal.
\newblock Sparse subspace clustering.
\newblock In \emph{2009 IEEE conference on computer vision and pattern recognition (CVPR)}, volume~6, pages 2790--2797, 2009.

\bibitem[Vidal et~al.(2003)Vidal, Soatto, Ma, and Sastry]{sc_st1}
R.~Vidal, S.~Soatto, Y.~Ma, and S.~Sastry.
\newblock An algebraic geometric approach to the identification of a class of linear hybrid systems.
\newblock In \emph{42nd IEEE international conference on decision and control (IEEE Cat. No. 03CH37475)}, volume~1, pages 167--172. IEEE, 2003.

\bibitem[Vidal et~al.(2008)Vidal, Tron, and Hartley]{sc_cv2}
R.~Vidal, R.~Tron, and R.~Hartley.
\newblock Multiframe motion segmentation with missing data using powerfactorization and gpca.
\newblock \emph{International Journal of Computer Vision}, 79:\penalty0 85--105, 2008.

\bibitem[Wang et~al.(2023)Wang, Sun, Yuan, Gao, and Lu]{multitvc}
L.~Wang, D.~Sun, Z.~Yuan, Q.~Gao, and Y.~Lu.
\newblock Multi-view clustering based on graph learning and view diversity learning.
\newblock \emph{The Visual Computer}, 39\penalty0 (12):\penalty0 6133--6149, 2023.

\bibitem[Wu et~al.(2018)Wu, Xiong, Yu, and Lin]{intsdisc}
Z.~Wu, Y.~Xiong, S.~X. Yu, and D.~Lin.
\newblock Unsupervised feature learning via non-parametric instance discrimination.
\newblock In \emph{Proceedings of the IEEE conference on computer vision and pattern recognition}, pages 3733--3742, 2018.

\bibitem[Yang et~al.(2008)Yang, Wright, Ma, and Sastry]{sc_cv1}
A.~Y. Yang, J.~Wright, Y.~Ma, and S.~S. Sastry.
\newblock Unsupervised segmentation of natural images via lossy data compression.
\newblock \emph{Computer Vision and Image Understanding}, 110\penalty0 (2):\penalty0 212--225, 2008.

\bibitem[Zhang et~al.(2019)Zhang, Li, You, Qi, Zhang, Guo, and Lin]{cnnSC}
J.~Zhang, C.-G. Li, C.~You, X.~Qi, H.~Zhang, J.~Guo, and Z.~Lin.
\newblock Self-supervised convolutional subspace clustering network.
\newblock In \emph{Proceedings of the IEEE/CVF conference on computer vision and pattern recognition}, pages 5473--5482, 2019.

\bibitem[Zhou et~al.(2019)Zhou, Xiao, Liu, Zhou, Hancock, et~al.]{DPSC_2019}
L.~Zhou, B.~Xiao, X.~Liu, J.~Zhou, E.~R. Hancock, et~al.
\newblock Latent distribution preserving deep subspace clustering.
\newblock In \emph{28th International joint conference on artificial intelligence}, 2019.

\end{thebibliography}

\clearpage
\appendix
\section{Architectures}
\begin{figure}[h]
    \centering
    \begin{subfigure}[b]{0.3\textwidth}
        \centering
        \includegraphics[width=\textwidth]{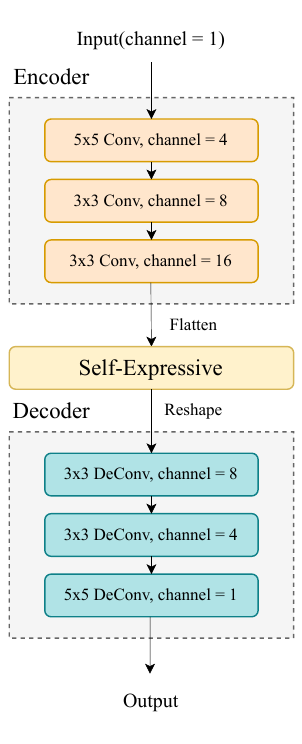}
        \caption{}
    \end{subfigure}
    \hfill
    \begin{subfigure}[b]{0.3\textwidth}
        \centering
        \includegraphics[width=\textwidth]{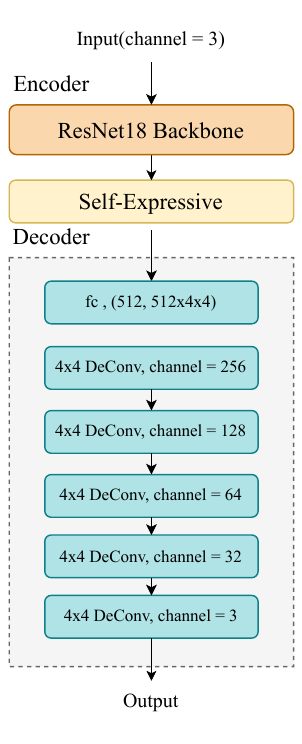}
        \caption{}
    \end{subfigure}
    \hfill
    \begin{subfigure}[b]{0.3\textwidth}
        \centering
        \includegraphics[width=\textwidth]{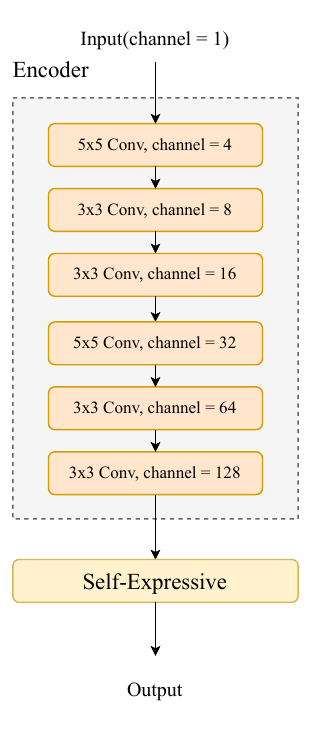}
        \caption{}
    \end{subfigure}
    \caption{Architectures for (a) BDSC on ORL, (b) BDSC on COIL100, and (c) CLBDSC on ORL.}
    \label{orl_coil100}
\end{figure}

Figure \ref{orl_coil100} presents the architectures used in Section \ref{sec_deeper}. All parts are randomly initialized except for the ResNet18 backbone. Since ORL is a grayscale dataset, the input channel of its networks is 1 instead of 3. 

As the self-expressive layer takes inputs of shape $(N, c*d)$, we flatten the outputs of encoders, then reshape the outputs of the self-expressive layer to $(N, c, \sqrt{d}, \sqrt{d})$ before feeding them into decoders. However, for BDSC on COIL100, since features are processed by average pooling and then flattened within the ResNet18 backbone, we employ a fully connected layer in the decoder to adjust the dimension to $(512, 4, 4)$. 

\end{document}